\documentclass[sigconf]{acmart}
\usepackage{booktabs}
\usepackage{epsfig}
\usepackage{epstopdf}
\usepackage{amssymb,amsmath}
\usepackage{amsmath}
\usepackage{graphicx}
\usepackage{caption}
\usepackage{graphicx} 

\setcopyright{rightsretained}
\acmConference{ICTIR' 17 Workshop on Search-Oriented Conversational AI (SCAI' 2017)}{October 1, 2017}{Amsterdam, Netherlands}

\begin{document}
\title{Deep Reinforcement Learning for Conversational AI}

\author{Mahipal Jadeja}

\affiliation{%
  \institution{DA-IICT}
  \city{Gandhinagar} 
  \state{India} }
\email{mahipaljadeja5@gmail.com}

\author{Neelanshi Varia}
\affiliation{%
  \institution{DA-IICT}
  \city{Gandhinagar} 
  \state{India} }
\email{neelanshiV2@gmail.com}

\author{Agam Shah}
\affiliation{%
	\institution{DA-IICT}
	\city{Gandhinagar} 
	\state{India} }
\email{shahagam4@gmail.com}


\begin{abstract}
Deep reinforcement learning is revolutionizing the artificial intelligence field. Currently,  it serves as a good starting point for constructing intelligent autonomous systems  which offer a better knowledge of the visual world. It is possible to scale deep reinforcement learning with the use of deep learning and do amazing tasks such as use of pixels in playing video games. In this paper, key concepts of deep reinforcement learning including reward function, differences between reinforcement learning and supervised learning and models for implementation of reinforcement are discussed. Key challenges related to the implementation  of reinforcement learning in conversational AI domain are identified as well as discussed in detail. Various conversational models which are based on deep reinforcement learning (as well as deep learning) are also discussed. In summary, this paper discusses key aspects of deep reinforcement learning which are crucial for designing an efficient conversational AI. 
\end{abstract}

%
%
%



\maketitle

\section{Introduction}
Artificial intelligence is playing role everywhere - banking, education, healthcare, services and almost every important sector. 
One of the key reason behind its success is conversational AI which has not only led us from typing commands to speaking while we are doing some other activity but also given us personal assistants which are almost humanlike in their speeches. Conversational AI will help us solve problems like language formation, context sensitive conversations, translation, better identification and other aspects which make the intelligent assistants more human-like. We have at our hand natural language processing, speech recognition, machine learning, neural networks, deep learning and other domains to transform the way we perceive artificial intelligence.


Deep/Hierarchical Learning is a subset of machine learning. It includes various architectures and neural networks which work on the information given to it. It works on the principle of knowledge building. It also predicts or classifies whether the knowledge is relevant or falls into which category. Reinforcement learning is one the three - supervised, unsupervised and reinforcement learning which is able to train a network by means of trial and error that is by punishing for error and rewarding for correct results. The deep reinforcement learning branch has emerged from the notion of training an artificially intelligent agent like human that is, to give it knowledge and improve by rewarding or punishing. A lot of research has already proved to better and better from what we had been seeing till years and it is expected that it will be one of the cornerstones for the dream future of AI.

\section{Deep Reinforcement Learning in Conversational AI}

Computers that can play games have always impressed the computing world. For computing world, computer machines that can play games excellently is always a topic of interest.  In a breakthrough paper published by DeepMind (London based company), with the use of Deep reinforcement learning, automated Atari playing $[1]$ was demonstrated.  After around one month of this amazing work, the company DeepMind was bought by Google.  After Google's entry in this field, there is a lot of buzz about reinforcement learning in the field of AI. 
A relatively recent success by Google is AlphaGo $[2]$ (artificial agent) who has won against the Go champion of the world. 

\begin{figure}[h]
	\includegraphics[scale=0.35]{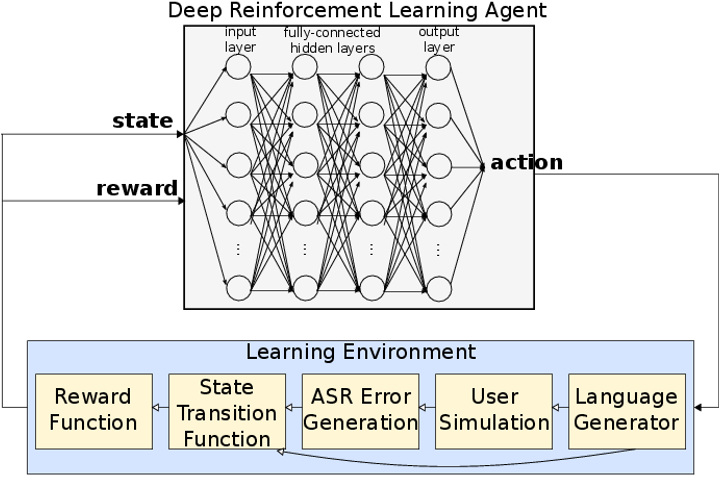}
	\caption{Deep Reinforcement Learning Agent (\ref{si1}) }
	\label{AR}
\end{figure}

\noindent \textbf {Basics of Reinforcement Learning}\\
Reinforcement learning is related to three broad fields namely $1)$ Artificial Intelligence, $2)$ animal psychology and $3)$ control theory. 
The idea is to have a robot/person/animal/deep net who is trying to learn to navigate in an environment which is dynamic and uncertain. The goal of the autonomous agent is to maximize a reward (see Figure~\ref{AR}) and generally, this reward is a quantitative entity (numeric).

It is easy to understand this concept with the help of sports.  In the case of Tennis, we can think about following actions of the virtual agent: serves, returns, volleys.  The state of the game depends upon the smart selection of these actions.  Here, the goal is to perform series of actions in order to win a point, game, set as well as match.  So this numeric reward is always being considered by the virtual agent.  The objective of the agent is to implement a strategy or a set of strategies in order to get best possible score. In other words, the objective is to maximize the scoring function of the game. 
Here the issue is: the state of the game is not static. Depending upon the actions of agents the state will change rapidly which makes such type of modeling very tricky. Input for this type of model is the present state of the game as well as an action and it is supposed to generate the best possible value for scoring function as an output.  But this scheme is just for one step whereas the overall objective is to win the game. Therefore the agent has to consider all the actions from the current state to the possible final state.  Therefore, this modeling approach is highly application dependent since for each application the scoring function will be different.  So one cannot use the same strategy which is used in building of Tennis agent for Chess agent and vice versa. 

 \footnote {\label {si1}  Source: https://ai2-s2-public.s3.amazonaws.com/figures/2016-11-08/ec4a764e062153c911097495c7e4b7e93612b75d/2-Figure1-1.png}

\noindent \textbf {Early Models of DRL}\\
In the case of Atari agent, convolutional neural network (with a lot of adjustments) was build by researchers with the help of Atari screenshots.  The scoring function was dependent upon a target number (maximum possible reward) not a class. 


Another model is Deep Q-Network which is also known by DQN and again this is a contribution from Google $[3]$.  It uses the same underlying principle: to maximize the reward points with the given state and action.  DQN offers improvements including but not limited to Experience Relay, Dueling Network Architecture. 

\noindent \textbf{Reinforcement learning vs Supervised Learning} \\
Reinforcement learning is not at all rewording of supervised learning.  In the case of supervised learning, the historical examples are used in order to understand the environment but this approach does not necessarily the best.  Consider the example of a car driving in heavy traffic to understand differences.  In the case of supervised learning, the idea is to use the past data (let's say 2 weeks before) in order to establish road patterns and use those patterns in the current scenario. But here the problem is it may possible that 2 weeks before, the roads were very clear in terms of traffic and today in a heavy traffic scenario, the available information is not that much useful and there is no effective way to use it in order to obtain best results.  Whereas, in the case of reinforcement learning, the focus is on rewards. The driver will get points for his/her action. Actions like maintaining speed of the vehicle less than the speed limit, lane driving, proper signaling as and when required etc.  Negative points are given for undesirable actions like speeding and tailgating. Here the objective function is to maximize the points and input is the current state of the traffic and action.  
Reinforcement learning focuses more on a change of the current state of the environment after each action and supervised learning models don't consider it. 

{ \footnote{\label {si2}   Source: https://qph.ec.quoracdn.net/main-qimg-b135e50fd568eac846f112ee8a0a1bbc } 
\begin{figure}[h]
	\includegraphics[scale=0.40]{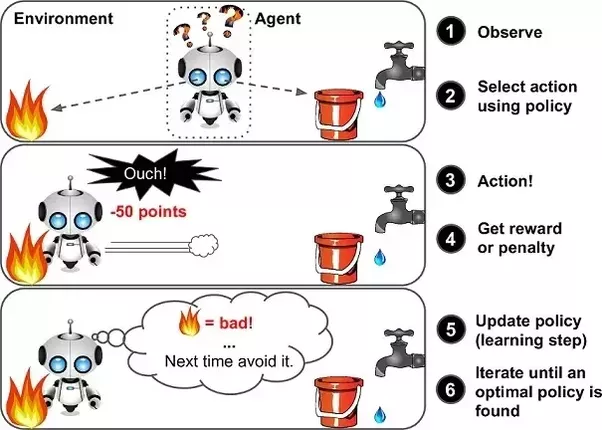}
	\caption{Summary of Reinforcement Learning (\ref{si2})}
	\label{RDLs}
\end{figure} 

\subsection{Study of Challenges in Reinforcement Learning in Conversational AI Domain}

Detailed study of reinforcement learning is beyond the scope of this paper.  So, instead of discussing mathematical equations or algorithms, we focus on challenges associated with reinforcement problems.

\noindent \textbf {Background: Reward Functions}

The reward functions provide a signal/feedback which is an indicator of the performance of the system with respect to the underlying action.  They also indicate the importance of each action by considering value addition by each action towards achieving the final goal/solution.   Supervised learning actually indicates which type of action is correct and should be taken by user whereas in the case of reinforcement learning, the only signal is given depending upon how good/bad the action is for achieving overall goal. There is no notion of correctness of local actions. 

There are two different ways to define reward functions for conversational AI: 1) Sparse functions  2) Non-sparse functions. Sparse functions are easy to design/define but very hard to solve whereas non-sparse functions are difficult to design but very easy to solve. In the case of sparse functions there is no signaling mechanism i.e. the user won't get feedback for his local choices and at terminal stage only, he/she will get information about whether the desired output is achieved or not. For example, consider sparse function defined for playing chess, where no feedback is given for local actions (moves) but towards the end of the game the user gets information about whether he/she has achieved the goal (winning the game) or not. In the case of a non-sparse function, depending on the usefulness of the function in achieving the final desired objective, signals are provided to the user. So that the user can drive the system for achieving an overall optimal solution. \\

We can make an analogy between the success of companies with rewards. Traditional conventional approach for most of the companies is to achieve finite limited rewards (profits) with known odds whereas other companies like Amazon wants to achieve out-sized massive rewards at long odds. The latter type of companies prefer exploration of new possibilities.  In the case of reinforcement learning, the idea is to select one path which gives the maximum value of expected reward by exploring trade-off between exploitation and exploration.  It may possible that a company gets a massive success after a long string of failures and the same thing is possible for rewards too. Therefore one can't ignore exploration part. Summary of reinforcement learning model is shown in Figure~\ref{RDLs}. \\
\textbf{Key Challenges:} \\

\noindent \textbf {Challenge 1: Multiple goals in the case of conversational AI} 
There are several objectives of a conversational AI including 1) Robust performance 2) Meaningful /informative interaction with the user 3) Provide excellent user experience 4) Offer personalization. 

So naturally, in the case of conversational AI, single reward function is not sufficient. The next challenge is how to assign weights to these goals/objectives? i.e. how much importance should be given to each of the desired objectives. In summary,  it is hard to design reward function which include these many challenges with appropriate weights.
\\
\noindent \textbf {Challenge 2: Trade-off between various goals} \\
The next challenge is to handle trade-off between the goals. For example,  it is difficult to offer extreme personalization as well as efficient performance for all the messages for a conversational AI.  So how to achieve optimal behavior in this scenario? 

According to us, designing a weighting scheme in order to combine several goals is the biggest challenge for conversational AI since most of the goals of conversational AI are depending upon users' experience which is difficult to quantify. Some type of automated negotiation between different goals is desirable using which it may possible to combine several objectives in a single way(action). But again, trading between different goals while considering the underlying environment is a very hard task. In the conversational AI there is also trade-off in generating dialogue between 1) length of dialogue 2) Diversity of dialogue and again 3) personalization.


\noindent \textbf {Challenge 3: Coherent dialogue design}\\
The agent should generate consistent response while generating answer for semantically identical input. For example, if a user ask question like "Where do you live now?" and "In which country do you live now?" he/she wants the same answer in both the cases. This problem looks simple but it is difficult to implement since the underling model should also generate linguistic plausible answer. Here, training data is huge and it consists data from multiple different users. A Persona-Based Neural Conversation Model is making first steps into the direction of explicitly modelling a personality $[6]$.

\noindent \textbf{Challenge 4: Evaluate conversational agents}\\
We can evaluate conversational agent by both subjective and objective evaluation technique. The subjective evaluation technique considers users' experiences of different aspects of the conversations, while the objective evaluation technique are based on an analysis of the logs of the actual conversations. These evaluation methods are well described in the literature $[11]$. Reward function can send feedback to agent based on this evaluation.

Evaluation of conversational agents depends upon quantitative as well
as qualitative features and most of the qualitative features are user
dependent. Therefore, we feel that extreme personalization and
universal defined metrics for qualitative features are the biggest
challenges for evaluation of conversational agents.

\subsection{Deep Reinforcement Learning based Conversational Models}
Deep reinforcement learning is  emerging area for development of conversational models $[7]$. Idea is to learn conversational pattern via trial and error method. Such training is performed via clients or a dialogue set predefined in computers. A huge dataset is required to train the deep neural network and so automatic chatbot algorithms are applied for training. For providing such training Bayesian models, Markov models, etc. have been developed. It is a great challenge to be able to model such algorithms which are accurate enough to train the reinforcement networks which includes gathering of relevant and sometimes specifically irrelavant dataset, semantics,etc. While following a statistical approach. To model dynamic training algorithm, human clients with enough knowledge and clarity of purpose or intelligent enough AI devices/algorithms are required which is a problem we face currently.

The above mentioned sequence to sequence model is able to generate dialogues given a conversation and context pre hand based on maximum likelihood estimation but it generates a very high amount of responses which in a way means that the intelligent agent is unable to answer. This model works on reward and punishment strategy like any other reinforcement model unlike MLE which helps in building long conversational training and learning for the AI assistant. Supervised learning in AlphaGo style strategy and optimisation techniques are also applied for the achievement.

By using large data and computing resources, the rise of deep reinforcement learning has boosted our ability to build computational models which are applicable in our lives. The AI bots built with Deep Reinforcement Learning understand the semantics of all domains and are capable of scaling. This advancement allows us to solve dialogue problems in various domains. The behaviour based on random, rule-based and supervision based learning outperformed by DRL based learned policies. A report of experiments concludes that the DRL-based policy has a $53\%$ win rate versus $3$ automated players, whereas a supervised player trained on a dialogue corpus in this setting achieved only $27\%$, versus the same $3$ bots $[10]$. The above results prove that DRL is a reliable framework for training dialogue systems and strategic agents with negotiation abilities. The experimental results report that all DRL agents substantially outperform all the baseline agents. 

\subsection{Deep Learning based Conversational Models}
Deep learning along with other emerging areas has greatly impacted the way we perceive artificial intelligence. With context to development of conversational artificial intelligence, deep learning has been able to make great leaps.  Evolution of deep learning $[9]$ is shown in Figure~\ref{AIf}.

Various aspects pertaining to speech and conversations have been addressed in the past and much work has to be done in the areas specific to conversational AI, speech recognition and natural language processing. This section addresses various models developed, comparison and further scope of research and challenges.

Neural networks and deep learning  help in user (model) based learning. The present intelligent personal assistant models are able to address any conversation at hand briefly. Recurrent Neural Network models $[4]$ are able to address 'intention' apart from 'attention'.  Origin RNN encodes the inputs so as to make the conversation more intentional and continuable whereas the destination neural network is able pay attention to specific words of user so as to keep learn from them and reply accordingly. The model is divided into three parts - first one collects the words and sentences spoken/entered by the user, second one captures the context of the conversation by various parts of sentence and third one saves various characteristics, objects, etc. This model is one step towards the intelligent assistant becoming more human-like and having conversations which are contextual and not absolute.  


The conversational AI is not limited to Siri, Cortana, Ok Google, etc. but is used in various mobile and web applications in the domains of healthcare, education, banking, etc. Some of these models are very specific based on their real-time application. The data fed in and the algorithm applied determines the efficiency of the personal assistant. One such important leap is in the education field. Smart boards and computers are limited in performance in the sense it depends on how the student accesses various materials. Auto-tutor is a conversational AI based model which teaches concepts via dialogues/conversations. Instead of providing information directly, it builds the knowledge based on questions posed by the agent as well as student. The tutor $[5]$. also possesses expressions, gestures, dialogues in natural language, etc. 

\footnote {\label{si3} Source: https://image.slidesharecdn.com/deeplearningframeworksslides-160531204623/95/deep-learning-frameworks-slides-6-638.jpg?cb=1464727714}
\begin{figure}[h]
	\includegraphics[scale=0.4]{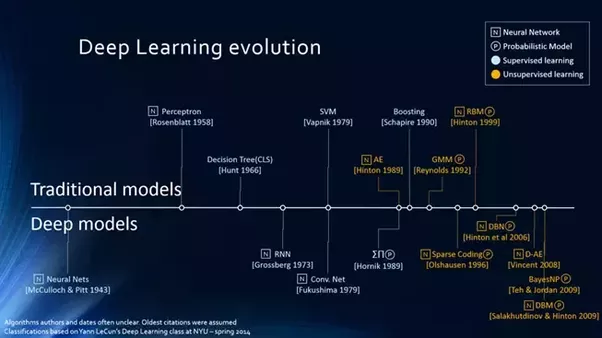}
	\caption{Deep learning evolution (\ref{si3})}
	\label{AIf}
\end{figure} 

Another aspect that is important is to take care of multi-users. That is, it might be possible that device is used by various clients for a particular use. In that case, the intelligent assistant should be able to serve according to the speaker. A neural model based on characteristics of the speaker $[6]$ has been developed. It takes care of output based on speaker, their characteristics, speaking methods, language and background knowledge. A speaker model is generated based on inputs and that records the features required to impersonate a model which has humanlike behaviour. Each individual speaker is considered as a vector that helps in encoding details regarding that particular user. Further, what we need is to take care of user data leakage and tampering by users. Speech recognition along with deep neural networks play an important role for development of these methods.

Deep Neural Networks (DNNs) are also another set of algorithms based on which dialogue generation and simulation models have been developed. The LSTM (Long-Short Term Memory) based approach vectors a dialogue system. It then includes various layers of vectored architectures to get the output.
 DNNs are able to perform parallel computing in a very optimised way which thus makes dialogue generation easier. But only those datasets whose inputs and outputs have a fixed dimension can be modelled which is a big limitation and working towards this has to be one of our future goals via sequential learning algorithms $[8]$, feedforward neural networks, recurrent neural networks, etc. 

\section{Conclusions} 
Reinforcement learning is the sub domain of artificial intelligence and it focuses on aspects like perception, goal setting and planing.  Reinforcement learning has potential for combining AI with other engineering disciplines. We conclude that reinforcement learning is simple yet powerful technique and it has a tremendous potential to contribute in the advancement of conversational based AI. For conversational AI, most of the challenges related to reinforcement learning are related to reward functions and therefore how to quantify user experiences/personalization in terms of reward function is one of the future direction of research. Since there are multiple goals in the case of conversational AI, the equally critical question is how to handle trade-off between various goals.

As seen in Section $2$, various mentioned models have their own achieved results and limitations. Combining them together to obtain all functionalities in one set to obtain a near perfect intelligent assistant is a future direction to work on. Individually, speech recognition, NLP, NN, Deep Learning, etc. have excelled in the field of producing conversational intelligence but to combine them to overcome limitations of different areas is where we should start working. Limitations consist of building a dataset, setting up the context of convention, detecting a speaker, performing a particular task and most importantly replying in a humanlike manner which includes features like natural language, sentence formation and translation, continuable reply.

\section{References}
\begin{enumerate}
	\item Mnih, V., Kavukcuoglu, K., Silver, D., Graves, A., Antonoglou, I., Wierstra, D., \& Riedmiller, M. (2013). Playing atari with deep reinforcement learning. arXiv preprint arXiv:1312.5602.
	\item   Chen, J. X. (2016). The evolution of computing: AlphaGo. Computing in Science \& Engineering, 18(4), 4-7.
	\item Van Hasselt, H., Guez, A., \& Silver, D. (2016, February). Deep Reinforcement Learning with Double Q-Learning. In AAAI (pp. 2094-2100).
	\item Yao, K., Zweig, G., \& Peng, B. (2015). Attention with intention for a neural network conversation model. arXiv preprint arXiv:1510.08565.
	\item Graesser, A. C., VanLehn, K., Rose, C. P., Jordan, P. W., \& Harter, D. (2001). Intelligent tutoring systems with conversational dialogue. AI magazine, 22(4), 39.
	\item Li, J., Galley, M., Brockett, C., Spithourakis, G. P., Gao, J., \& Dolan, B. (2016). A persona-based neural conversation model. arXiv preprint arXiv:1603.06155.
	\item Li, J., Monroe, W., Ritter, A., Galley, M., Gao, J., \& Jurafsky, D. (2016). Deep reinforcement learning for dialogue generation. arXiv preprint arXiv:1606.01541.
	\item Sutskever, I., Vinyals, O., \& Le, Q. V. (2014). Sequence to sequence learning with neural networks. In Advances in neural information processing systems (pp. 3104-3112).
	\item LeCun, Y., Bengio, Y., \& Hinton, G. (2015). Deep learning. Nature, 521(7553), 436-444.
	\item Cuayahuitl, H., Keizer, S., \& Lemon, O. (2015). Strategic dialogue management via deep reinforcement learning. arXiv preprint arXiv:1511.08099.
	\item Silvervarg, A., \& Jonsson, A. (2011, July). Subjective and objective evaluation of conversational agents in learning environments for young teenagers. In Proceedings of the 7th IJCAI Workshop on Knowledge and Reasoning in Practical Dialogue Systems.
\end{enumerate}

\end{document}